\pdfoutput=1

\documentclass[11pt,a4paper]{article}
\usepackage[hyperref]{acl2021}
\usepackage{times}
\usepackage{latexsym}

\usepackage{microtype}

\aclfinalcopy %
\def\aclpaperid{308} %

\usepackage{amsmath,amsfonts,bm}
\usepackage{booktabs} %
\usepackage{url}
\usepackage{graphicx}
\usepackage{array}
\usepackage{color}
\usepackage{makecell}
\usepackage{multirow}
\usepackage{xspace}

\newcommand{\ie}{\textit{i.e.}} %
\newcommand{\eg}{\textit{e.g.}} %
\newcommand{\start}[1]{\vspace{.7mm}\noindent{{\bf #1}.}}

\newcommand{\upv}{\vspace{-.0cm}}
\newcommand{\downv}{\vspace{-.0cm}}

\newcommand{\ours}{\textsc{Rider}\xspace}
\newcommand{\GAR}{\textsc{Gar}\xspace}
\newcommand{\GARPlus}{$\textsc{Gar}^{\texttt{+}}$\xspace}

\definecolor{gred}{RGB}{219,68,55}
\definecolor{gblue}{RGB}{66,133,244}
\definecolor{gyellow}{RGB}{244,180,0}
\definecolor{ggreen}{RGB}{15,157,88}
\definecolor{ggrey}{RGB}{115,115,115}

\title{\ours: Reader-Guided Passage Reranking \\ for Open-Domain Question Answering}

\author{\makecell{Yuning Mao$^{1}$\thanks{\hspace{.06in}Work was done during internship at Microsoft Azure AI.}, Pengcheng He$^{2}$, Xiaodong Liu$^{3}$, Yelong Shen$^2$,\\ Jianfeng Gao$^{3}$, Jiawei Han$^1$, Weizhu Chen$^2$} \\
$^1$University of Illinois, Urbana-Champaign \quad
$^2$Microsoft Azure AI \quad
$^3$Microsoft Research\\
$^1$\{yuningm2, hanj\}@illinois.edu \\ $^{2,3}$\{penhe, xiaodl, yeshe, jfgao,wzchen\}@microsoft.com 
} 

\date{}

\begin{document}
\maketitle
\begin{abstract}
Current open-domain question answering systems often follow a Retriever-Reader architecture, where the retriever first retrieves relevant passages and the reader then reads the retrieved passages to form an answer.
In this paper, we propose a simple and effective passage reranking method, named Reader-guIDEd Reranker (\ours), which does not involve training and reranks the retrieved passages solely based on the top predictions of the reader before reranking.
We show that \ours, despite its simplicity, achieves 10 to 20 absolute gains in top-1 retrieval accuracy and 1 to 4 Exact Match (EM) gains without refining the retriever or reader.
In addition, \ours, without any training, outperforms state-of-the-art transformer-based supervised rerankers.
Remarkably, \ours achieves 48.3 EM on the Natural Questions dataset and 66.4 EM on the TriviaQA dataset when only 1,024 tokens (7.8 passages on average) are used as the reader input after passage reranking.\footnote{Our code is available at \url{https://github.com/morningmoni/GAR}.}
\end{abstract}

\section{Introduction}
Current open-domain question answering (OpenQA) systems often follow a Retriever-Reader (R2) architecture, where the retriever first retrieves relevant passages and the reader then reads the retrieved passages to form an answer.
Since the retriever retrieves passages from a large candidate pool (\eg, millions of Wikipedia passages), it often fails to rank the most relevant passages at the very top.
One line of work \cite{mao2020generation,karpukhin2020dense} aims to improve the retriever and shows that significantly better QA performance can be achieved when the retrieval results are improved.

An alternative solution is to rerank the initial retrieval results via a reranker, which is widely used in information retrieval \cite{nogueira2019passage,qiao2019understanding} and explored in early OpenQA systems \cite{wang2017r,lee-etal-2018-ranking}.
However, current state-of-the-art OpenQA systems \cite{karpukhin2020dense,izacard2020leveraging} do not distinguish the order of the retrieved passages and instead equally consider a large number of retrieved passages (\eg, 100), which could be computationally prohibitive as the model size of the readers becomes larger \cite{izacard2020leveraging}.

We argue that a Retriever-Reranker-Reader (R3) architecture is beneficial in terms of both model effectiveness and efficiency: passage reranking improves the retrieval accuracy of the retriever at top positions and allows the reader to achieve comparable performance with fewer passages as the input.
However, one bottleneck of R3 is that the reranker, previously based on BiLSTM \cite{wang2017r,lee-etal-2018-ranking} and nowadays typically BERT-based cross-encoder \cite{nogueira2019passage,qiao2019understanding}, is often costly to train and its slow inference delays the whole pipeline as well.

Can we achieve better performance without the bother of training an expensive reranker or refining the retriever (reader)?
In this paper, we propose a simple and effective passage reranking method, named \textbf{R}eader-gu\textbf{IDE}d \textbf{R}eranker (\ours), which does not require any training and reranks the retrieved passages solely based on their lexical overlap with the top predicted answers of the reader before reranking.
Intuitively, the top predictions of the reader are closely related to the ground-truth answer and even if the predicted answers are partially correct or incorrect, they may still provide useful signals suggesting which passages may contain the correct answer \cite{mao2020generation}.

We conduct experiments on the Natural Questions (NQ) \cite{kwiatkowski-etal-2019-natural} and TriviaQA (Trivia) \cite{joshi-etal-2017-triviaqa} datasets.
We demonstrate that R3 with \ours, without any additional training, achieves 10 to 20 absolute gains in top-1 retrieval accuracy, and 1 to 4 gains in Exact Match (EM) compared to the R2 architecture.
\ours also outperforms two state-of-the-art transformer-based supervised reranking models that require expensive training and inference.
Notably, using only 1,024 tokens (7.8 passages on average) as the input of a generative reader, \ours achieves EM=47.5/63.5 on NQ/Trivia when the predictions of the same generative reader (EM=45.3/62.2 in R2)  are used for reranking, and EM=48.3/66.4 on NQ/Trivia when the predictions of an extractive reader (EM=43.8/62.7 in R2) are used for reranking.

\start{Contributions}
(1) We propose Reader-guIDEd Reranker (\ours), a simple and effective passage reranking method for OpenQA, which reranks the retriever results by the reader predictions without additional training and can be easily applied to existing R2 systems for performance improvements.
(2) We demonstrate that the passages reranked by \ours achieve significantly better retrieval accuracy and consequently lead to better QA performance without refining the retriever or reader.
(3) Notably, \ours achieves comparable or better performance than state-of-the-art methods on two benchmark datasets when only 1,024 tokens are used as the reader input after passage reranking.

\section{Method}
\subsection{Task Formulation}
We assume that an OpenQA system with an R2 architecture is available.
We denote the initially retrieved passages of the retriever as $\mathbf{R}$. We denote the top-N predictions of the reader on the top-k passages of $\mathbf{R}$ (denoted as $\mathbf{R}^{[:k]}$) as $\mathbf{A}^{[:N]}$.
The goal of \ours is to rerank $\mathbf{R}$ to $\mathbf{R'}$ using $\mathbf{A}^{[:N]}$ such that the retrieval accuracy is improved and better end-to-end QA results are achieved when $\mathbf{R'}^{[:k]}$ is used as the reader input instead of $\mathbf{R}^{[:k]}$.

\subsection{Passage Reranking}
Given an initially retrieved passage list $\mathbf{R}$ and top-N predictions of the reader $\mathbf{A}^{[:N]}$, \ours forms a reranked passage list $\mathbf{R'}$ as follows.
\ours scans $\mathbf{R}$ from the beginning of the list and appends to $\mathbf{R'}$ every passage $p \in \mathbf{R}$ if $p$ contains any reader prediction $a \in \mathbf{A}^{[:N]}$ after string normalization (removing articles and punctuation) and tokenization. Then, the remaining passages are appended to $\mathbf{R'}$ according to their original order.

Intuitively, if the reader prediction is perfect, the retrieval accuracy after reranking is guaranteed to be optimal.
Specifically, if the reader prediction is correct, it is guaranteed that the retrieval accuracy after reranking is better, since \ours moves all passages containing the correct answer to the top (or at least the same if those passages are \textit{all} at the top before reranking). If the reader prediction is wrong, \ours could still be better if the predicted answer co-occurs with the correct answer, the same, or worse if the predicted answer is misleading. 
In practice, if the reader performs reasonably well, \ours is also likely to rerank passages well.
Overall, we observe quantitatively that \ours leads to consistent gains in terms of both retrieval accuracy and QA performance without refining the retriever (reader) or even any training itself despite the noise in reader predictions.

\subsection{Passage Reading}
\label{sec:reading}
We consider a scenario where the number of passages that can be used for QA is limited (sometimes deliberately) due to reasons such as insufficient computational resources, the limit of model input length, or requirement for faster responses.
We use a generative reader initialized by BART-large \cite{lewis2019bart}, which concatenates the question and top-10 retrieved passages, trims them to 1,024 tokens (7.8 passages are left on average) as the input, and learns to generate the answer in a seq2seq manner \cite{mao2020generation,min2020ambigqa}.
We further add a simple shuffle strategy during reader training, which randomly shuffles the top retrieved passages before concatenation.
In this way, the reader appears to be more robust to the reranked passages during inference and achieves better performance after reranking.

\begin{table*}[t]
\centering

\resizebox{2\columnwidth}{!}{
\begin{tabular}{l ccccc | ccccc}
  \toprule
    \multirow{2}{*}{\textbf{Data Input}}  & \multicolumn{5}{c|}{\textbf{NQ}} & \multicolumn{5}{c}{\textbf{Trivia}} \\
    & Top-1 & Top-5 & Top-10 & Top-20 & Top-100 & Top-1 & Top-5 & Top-10 & Top-20 & Top-100 \\
\midrule
    $\mathbf{R}$        & 46.8 & 70.7& 77.0 & 81.5 & 88.9 &  53.2 & 73.1 & 77.0 & 80.4 & 85.7 \\ 
    $\mathbf{R'}$ by $\mathbf{G}$ (N=1)        & \textbf{58.6} & 71.4 & 76.9 & 81.6 & 88.9 & \textbf{68.8} & 74.8 & 77.5 & 80.4 & 85.7 \\
    $\mathbf{R'}$ by $\mathbf{G}$ (N=10)        & 56.4 & \textbf{72.2}& \textbf{77.3} & 81.6 & 88.9 & 66.9 & \textbf{75.3} & \textbf{77.9} & \textbf{80.8} & 85.7 \\
\midrule
    $\mathbf{R'}$ by $\mathbf{E}$ (N=1)        & \textbf{60.4} & 72.1 & 77.3 & 81.7 & 88.9 & \textbf{71.9} & 77.5 & 79.8 & 81.8 & 85.7 \\
    $\mathbf{R'}$ by $\mathbf{E}$ (N=5)     & 53.5 &\textbf{75.2} & \textbf{80.0} & 83.2 & 88.9 & 63.2 & \textbf{77.9} & \textbf{80.7} & 82.8 & 85.7 \\
    $\mathbf{R'}$ by $\mathbf{E}$ (N=10)     & 50.3 & 74.3 & 80.0 & \textbf{84.2} & 88.9 & 59.8 & 77.1 & 80.5 & \textbf{82.9} & 85.7  \\

  \bottomrule
\end{tabular}
}

\caption[Caption]{\textbf{Top-k retrieval accuracy on the test sets before ($\mathbf{R}$) and after ($\mathbf{R'}$) reranking}. $\mathbf{G}$ and $\mathbf{E}$ denote generative and extractive readers, respectively, whose top predictions are used for reranking.}
\label{tab:top_k_acc}
\end{table*}

\section{Experiment Setup}

\start{Datasets}
 We conduct experiments on the open-domain version of two widely used QA benchmarks -- Natural Questions (NQ) \cite{kwiatkowski-etal-2019-natural} and TriviaQA (Trivia) \cite{joshi-etal-2017-triviaqa}, whose statistics are listed in Table~\ref{tab:dataset}.

\begin{table}[ht]
\centering

\resizebox{\columnwidth}{!}{
\scalebox{1}{
\begin{tabular}{llrrr}
\toprule
\textbf{Dataset} & \textbf{Train / Val / Test} & \textbf{Q-len} & \textbf{A-len} & \textbf{\#-A}\\
\midrule
NQ &79,168 / 8,757 / 3,610	&12.5	& 5.2 & 1.2\\
Trivia                 &         78,785 / 8,837 / 11,313  &    20.2  &  5.5 &  13.7\\

\bottomrule
\end{tabular}
}
}
\upv
\caption{\textbf{Dataset statistics} that show the number of samples, the average question (answer) length, and the average number of answers for each question.
}
\label{tab:dataset}
\downv
\end{table}

\start{Evaluation Metrics}
Following prior studies \cite{mao2020generation,karpukhin2020dense}, we use top-k retrieval accuracy to evaluate the retriever and Exact Match (EM) to evaluate the reader.
\textit{Top-k retrieval accuracy} is the proportion of questions for which the top-k retrieved passages contain at least one answer span. It is an upper bound of how many questions are answerable by an extractive reader.
\textit{Exact Match (EM)} is the proportion of the predicted answer spans being exactly the same as one of the ground-truth answers, after string normalization such as article and punctuation removal.

\start{Source of $\mathbf{R}$}
Following \citet{mao2020generation}, we take the top 100 retrieved passages of \GAR~\cite{mao2020generation} on Trivia and its combination with DPR \cite{karpukhin2020dense} on NQ (\GARPlus) as the initial retrieval results $\mathbf{R}$ for reranking.

\start{Source of $\mathbf{A}^{[:N]}$}
To obtain the top-N predicted answers, we first take the predictions of the generative reader ($\mathbf{G}$) in Sec.~\ref{sec:reading}, which is trained on the passages without reranking and used for final passage reading in R3. It represents an apple-to-apple comparison to R2 without any additional information but higher-quality input.
We also experiment with an extractive reader ($\mathbf{E}$) that has access to all retrieved passages, where the goal is to study whether we can rerank passages via other signals and further improve $\mathbf{G}$ such that it outperforms both $\mathbf{G}$ and $\mathbf{E}$ when they are in R2.
We use the extractive reader in \citet{mao2020generation} with BERT-base \cite{devlin-etal-2019-bert} representation and span voting.

For the generative reader, we either take its top-1 prediction with greedy decoding or sample 10 answers with decoding parameters as follows.
We set sampling temperature to 5/2 and the top probability in nucleus sampling to 0.5/0.5 on NQ/Trivia, respectively.
Note that there are duplicate samples and on average $\bar{N}=6$.
We set the max input length to 1,024 and max output length to 10.
For the extractive reader, the top predictions are the text spans with the highest scores and we set $N=1, 5, 10$.

\section{Experiment Results}

\subsection{Quality of Reranking Signals}
We first analyze the EM of the top-N reader predictions $\mathbf{A}^{[:N]}$. 
We consider a question correctly answered as long as one of the top-N predictions matches the ground-truth answer. 
The standard EM is a special case with $N=1$.
As listed in Table~\ref{tab:top_n_em}, the reader EM can be improved by up to 24.0 on NQ and 15.8 on Trivia if we consider the top-10 predictions instead of only the first prediction, suggesting that there is significant potential if we use multiple predicted answers for reranking. That said, using more reader predictions also introduces more noise, \ie, incorrect answers, which could be misleading at times.

\begin{table}[ht]
\centering

\resizebox{\columnwidth}{!}{
\scalebox{1}{
\begin{tabular}{l cccc}
\toprule
\textbf{Dataset} & \textbf{Top-1} & \textbf{Top-3} & \textbf{Top-5} & \textbf{Top-10}\\
\midrule
NQ & 43.8 (45.3) 	& 57.4	& 62.6 & 67.8 (54.2)\\
Trivia  & 62.7 (62.2)  &    72.6  &  75.5 &  78.5 (67.7)\\

\bottomrule
\end{tabular}
}
}
\upv
\caption{\textbf{EM of top-N predictions of the reader.} Results are mostly on reader $\mathbf{E}$. Only top-1 and top-10 EM are shown (in the brackets) for reader $\mathbf{G}$, as its 10 predictions are sampled without particular order.
}
\label{tab:top_n_em}
\downv
\end{table}

\subsection{\ours for Passage Retrieval}
We list the top-k retrieval accuracy before and after passage reranking in Table~\ref{tab:top_k_acc}.
\ours significantly improves the retrieval accuracy at top positions (especially top-1) without refining the retriever.
In particular, we observe that when taking more reader predictions (\ie, larger N), the top-k retriever accuracy tends to improve more at a larger $k$ and less at a smaller $k$.
For example, an improvement of about 3 points is achieved for top-5 and top-10 accuracy when increasing $N$ from 1 to 5 on NQ for reader $\mathbf{E}$, but the top-1 retrieval accuracy also drops significantly (although still better than without reranking), which again suggests that there is a trade-off between answer coverage and noise.
Note that the top-100 retrieval accuracy is unchanged after reranking since we rerank the top-100 passages.

\subsection{\ours for Passage Reading}
\start{Comparison w. the state-of-the-art}
We show the QA performance comparison between \ours and state-of-the-art methods in Table~\ref{tab:sota}. We observe that \ours improves \GAR (or \GARPlus) on both datasets by a large margin, despite that they use the same generative reader and no further model training is conducted. Such results indicate that \ours provides higher-quality input for the reader and better performance can be achieved with the same input length. 
Moreover, the results of \ours are better than most of the existing methods that take more passages as input, except for FID-large~\citep{izacard2020leveraging} that reads 100 passages and also has more model parameters.

\begin{table}[ht]
\centering

\resizebox{1.0\columnwidth}{!}{
\begin{tabular}{clcc}
  \cmidrule[0.06em]{2-4}
    &\textbf{Method} & \textbf{NQ} & \textbf{Trivia} \\

\cmidrule{2-4}

  \multirow{9}{*}{\rotatebox[origin=c]{90}{Extractive}} &Hard EM~\citep{min-etal-2019-discrete}          & 28.1 & 50.9    \\
  &Path Retriever~\citep{asai2019learning}  & 32.6 & -     \\
  &ORQA~\citep{lee-etal-2019-latent}               & 33.3 & 45.0   \\
  &Graph Retriever~\citep{min2019knowledge} & 34.5 & 56.0   \\
  &REALM~\citep{guu2020realm}               & 40.4 & -      \\
  &DPR~\citep{karpukhin2020dense}           & 41.5 & 57.9 \\
  &BM25 \cite{mao2020generation} & 37.7 & 60.1 \\
  &\GAR & \textbf{41.8} & \textbf{62.7}  \\
  &\GARPlus & \textbf{43.8}& -  \\
  
\cmidrule{2-4}

\multirow{11}{*}{\rotatebox[origin=c]{90}{Generative}} &GPT-3~\citep{brown2020language}    & 29.9 & -   \\
    &T5~\citep{roberts2020much} & 36.6 & 60.5 \\
  
  &SpanSeqGen~\citep{min2020ambigqa}        & 42.2 & -      \\
  &RAG~\citep{lewis2020retrieval}           & 44.5 & 56.1  \\
  &FID-base \cite{izacard2020leveraging}    & 48.2 & 65.0 \\
  &FID-large \cite{izacard2020leveraging}   & \textbf{51.4} & \textbf{67.6}  \\
  &BM25 \cite{mao2020generation} & 35.3 & 58.6 \\
  \cmidrule{3-4}
  &\GAR \cite{mao2020generation}  & 38.1 & 62.2 \\
  &\textbf{\ours (\GAR)}  & - & \textbf{66.4} \\
  \cmidrule{3-4}
  &\GARPlus \cite{mao2020generation} & 45.3 & -  \\
  &\textbf{\ours (\GARPlus)} & \textbf{48.3} & -  \\

  \cmidrule[0.06em]{2-4}
\end{tabular}
}

\caption[Caption]{\textbf{End-to-end QA comparison of state-of-the-art methods}. \ours results in up to 4.2 EM gains.}
\label{tab:sota}
\end{table}

\start{Ablation Study}
A detailed analysis of \ours with different reranking signals is shown in Table~\ref{tab:rerank_comparison}. 
By reranking based on the prediction of the generative reader $\mathbf{G}$ (with input $\mathbf{R}^{[:k]}$), \ours generally achieves 1 to 2 gains in EM, which shows that \ours can improve end-to-end QA performance without any additional information.
By iterative reranking ($\mathbf{R''}$) using the reader predictions after first reranking, the performance of \ours is further improved. Conducting more than two iterations of reranking does not appear to bring additional gains.

\ours achieves even better performance when using the predictions of the extractive reader $\mathbf{E}$ (with input $\mathbf{R}$) for reranking, which is consistent with the results on retrieval. It is also encouraging to see that \ours significantly outperforms $\mathbf{E}$, which is more computationally expensive and has access to much more passages.

\begin{table}[t]
\centering

\resizebox{1.\columnwidth}{!}{
\begin{tabular}{lrr}
  \toprule
     \textbf{Data Input}  & \textbf{NQ} & \textbf{Trivia}  \\
\midrule
    $\mathbf{R}$  &  45.3 & 62.2 \\
    $\mathbf{R'}_1$ by $\mathbf{G}$ (N=1)  &  (45.3+1.1) 46.4  & (62.2+0.7) 62.9  \\
	$\mathbf{R'}_2$ by $\mathbf{G}$ (N=10)  &  (45.3+2.1) \textbf{47.4}  & (62.2+0.9) \textbf{63.1}  \\
	$\mathbf{R''}$ by $\mathbf{R'}_1$  &  (46.4+1.1) \textbf{47.5}  & (62.9+0.6) \textbf{63.5}  \\
	\midrule
	$\mathbf{R'}$ by $\mathbf{E}$ (N=1)  &  (43.8+3.2) 47.0  & (62.7+3.4) \textbf{66.1}  \\
	$\mathbf{R'}$ by $\mathbf{E}$ (N=5)  &  (43.8+4.5) \textbf{48.3}  & (62.7+2.5) 65.2  \\
	$\mathbf{R''}$ by best $\mathbf{R'}$  &  (48.3+0) \textbf{48.3}  & (66.1+0.3) \textbf{66.4}  \\

  \bottomrule
\end{tabular}
}

\caption[Caption]{\textbf{Comparison of \ours in EM when different reranking signals are used}. The numbers in the brackets represent the performance of the reader used for reranking and relative gains.}
\label{tab:rerank_comparison}
\end{table}

\begin{table}[t]
\centering

\resizebox{1\columnwidth}{!}{
\begin{tabular}{l ccccc}
  \toprule
    \textbf{Method} & \textbf{Top-1} & \textbf{Top-5} & \textbf{Top-10} & \textbf{Top-20}\\
\midrule
    $\mathbf{R}$        & 46.8 & 70.7& 77.0 & 81.5 \\ 
    $\mathbf{R'}$ by BERT reranker        & 51.4 & 67.6 & 75.7 & 82.4 \\
    $\mathbf{R'}$ by BART reranker        & 55.2 & 73.5 & 78.5 & 82.2 \\
    $\mathbf{R'}$ by $\mathbf{G}$ (N=10)        & \textbf{56.4} & 72.2& 77.3 & 81.6 \\
    $\mathbf{R'}$ by $\mathbf{E}$ (N=5)     & 53.5 &\textbf{75.2} & \textbf{80.0} & \textbf{83.2}  \\

  \bottomrule
\end{tabular}
}

\caption[Caption]{\textbf{Comparison with supervised rerankers in top-k retrieval accuracy on NQ}. \ours outperforms expensive transformer-based models without training.}
\label{tab:top_k_acc_supervised}
\end{table}

\subsection{Comparison w. Supervised Reranking}
Finally, we compare \ours with two state-of-the-art supervised reranking models. The first reranker is a BERT-base cross-encoder \cite{nogueira2019passage}, which is popularly used for passage reranking in information retrieval. The cross-encoder concatenates the query and passage, and makes a binary relevance decision for each query-passage pair. The second one generates relevance labels as target tokens in a seq2seq manner \cite{nogueira-etal-2020-document}. We use BART-large as the base model and ``YES/NO'' as the target tokens.

As listed in Table~\ref{tab:top_k_acc_supervised}, \ours, without any training, outperforms the two transformer-based supervised rerankers on retrieval accuracy.
Also, for QA performance, the best EM we obtain using the supervised rerankers is merely 46.3 on NQ.
Such results further demonstrate the effectiveness of \ours, which has the advantage of utilizing information from multiple passages (when the reader makes predictions), while the other rerankers consider query-passage pairs independently.

\subsection{Runtime Efficiency}
The reranking step of \ours only involves string processing, which can be easily paralleled and reduced to within seconds.
We use Nvidia V100 GPUs for reader training and inference.
The training of the generative reader takes 8 to 10 hours with 1 GPU, while it takes 12 to 16 hours with 8 GPUs for the DPR reader~\citep{karpukhin2020dense}.
Due to fewer input passages, the inference of the generative reader is also very efficient -- it takes around 3.5/11 min to generate answers on the NQ/Trivia test set with 1 GPU. In comparison, the DPR reader takes about 14/40 min with 8 GPUs.

\section{Related Work}

\start{Reranking for OpenQA}
Reranking has been widely used in information retrieval to refine the initial retrieval results.
Early effort on passage reranking for OpenQA uses supervised \cite{lee-etal-2018-ranking} or reinforcement learning~\cite{wang2017r} based on BiLSTM.
More recently, BERT-based rerankers that treat the query and passage as a sentence pair (\ie, cross-encoders) achieve superior performance \cite{nogueira2019passage,qiao2019understanding}.
However, the training of cross-encoders is rather costly. Moreover, the representations of cross-encoders cannot be pre-computed and matched via Maximum Inner Product Search (MIPS) as in bi-encoders \cite{karpukhin2020dense} but measured online between the query and each passage, which results in slower inference as well.

Another line of work \cite{das2018multi,qi2020retrieve} reranks the passages by updating the query and often involves a complicated learning process such as R2 interactions.
Alternatively, some prior studies \cite{wang2018evidence,iyer2020reconsider} directly rerank the top predicted answers instead of the passages using either simple heuristics or additional training.
In contrast, \ours utilizes downstream signals (\ie, the predictions of a reader) to rerank the passages without any training.

\start{Reader Distillation}
Recent studies~\cite{izacard2020distilling,yang2020retriever} show that distillation from the preference of the reader can improve the retriever performance, where the reader preference is measured by the attention scores of the reader over different passages and the retriever is refined by learning to approximate the scores.
\ours, to some extent, can also be seen as one way to distill the reader. However, \ours is much simpler in that no further training is involved for either the retriever or reader, and explicit reader predictions instead of latent attention scores are leveraged to improve the retriever results directly.

\section{Conclusion}
In this work, we propose \ours, a simple and effective passage reranking method for OpenQA, which does not involve additional training or computationally expensive inference, and outperforms state-of-the-art supervised rerankers that involve both.
\ours can be easily integrated into existing R2 systems for performance improvements.
Without fine-tuning the retriever or reader, \ours improves the retrieval accuracy and the QA results on two benchmark datasets significantly.
Notably, \ours achieves comparable or better performance than state-of-the-art methods with less reader input and allows for more efficient OpenQA systems. For future work, we will explore other simple and effective reranking strategies with no (minimal) training or external supervision.

\section*{Acknowledgments}
We thank Hao Cheng, Sewon Min, and Wenda Qiu for helpful discussions.
We thank the anonymous reviewers for valuable comments.

\bibliographystyle{acl_natbib}
\bibliography{anthology,main}

\end{document}